\documentclass{mva_style}
\usepackage{graphicx}
\usepackage{amsmath,booktabs}
\usepackage{amssymb}
\usepackage{diagbox}
\usepackage[top=22.5mm, bottom=22.5mm, left=22.5mm, right=22.5mm]{geometry}
\usepackage{multirow}
\usepackage{color, colortbl}
\usepackage[breaklinks=true,colorlinks,bookmarks=false]{hyperref}

\usepackage{url}
\finalcopy 

\begin{document}
\title{TomatoDIFF: On--plant Tomato Segmentation with Denoising Diffusion Models\thanks{This work was financed by the Slovenian Research Agency 
 (ARRS), research projects [J2-2506] and [J2-2501 (A)], research programmes [P2-0095] and [P2-0250 (B)] and by the UL Innovation Fund}}

\author{
  Marija Ivanovska, Vitomir \v{S}truc, Janez Per\v{s}\\
  Faculty of Electrical Engineering, University of Ljubljana\\
  Tr\v{z}a\v{s}ka cesta 25, Ljubljana, Slovenia\\
  {\tt \{marija.ivanovska, vitomir.struc, janez.pers\}@fe.uni-lj.si}\\
}

\maketitle

\section*{\centering Abstract}
\textit{
  Artificial intelligence applications enable farmers to optimize crop growth and production while reducing costs and environmental impact. Computer vision-based algorithms in particular, are commonly used for fruit segmentation, enabling in-depth analysis of the harvest quality and accurate yield estimation. In this paper, we propose TomatoDIFF, a novel diffusion-based model for semantic segmentation of on-plant tomatoes. When evaluated against other competitive methods, our model demonstrates state-of-the-art (SOTA) performance, even in challenging environments with highly occluded fruits. Additionally, we introduce Tomatopia, a new, large and challenging dataset of greenhouse tomatoes. The dataset comprises high-resolution RGB-D images and pixel-level annotations of the fruits. The source code of TomatoDIFF and Tomatopia are available at \url{https://github.com/MIvanovska/TomatoDIFF}.
}

\vspace{-2pt}
\section{Introduction}
\vspace{-1pt}
Tomatoes are among the most popular and widely consumed fruits, grown in large fields that require continuous monitoring~\cite{Tian_20201_cv_agriculture}. 
With the advent of Industry $4.0$, artificial intelligence (AI) has become a valuable tool for boosting production efficiency in food farms~\cite{Dhanya_2022_AI_agriculture}. Computer vision, in particular, is used for monitoring soil conditions~\cite{Shaik_2018_soil}, fruit quality assessment~\cite{rahim_2021_tomato_maturity}, detecting plant diseases~\cite{Shoaib_2022_leaf_disease}, predicting crop yield~\cite{Khaki_2020_crop_prediction}, and assessing fruit quality~\cite{rahim_2021_tomato_maturity}. Accurate segmentation and localization of objects of interest are crucial for in-depth visual analysis.


Lately, there has been a rapid development of automated computer vision applications for localization and segmentation of different tomato cultivars~\cite{Ojo_2022_recent_review}.
Classic algorithms often detect crops by color and shape~\cite{Malik2019_seg_loc_rgbd}. These traditional approaches however lack robustness in environments with variable lighting conditions or when the foliage is dense. Detection of green tomatoes is also challenging since their color can be barely distinguished from the surrounding vegetation. More recent algorithms take advantage of modern, deep learning models. Tomatoes specifically are often localized with single--stage algorithms such as Faster R-CNN, YOLO or SSD~\cite{Mu_2020_faster_Rcnn, Magalhaes_2021_SSD_YOLO_agRob}. Once the crops are detected, quiality estimation and yield prediction are performed by segmenting the fruits with various hand--crafted or neural network--based techniques~\cite{Zu_2021_maskrcnn_green_tomato}. Despite being much more accurate than classic approaches, frequently used deep learning methods often have difficulties with the detection of highly occluded fruits or crops that are further away from the camera.

\begin{figure}[t]
\begin{center}
\centering
  \includegraphics[width=\linewidth]{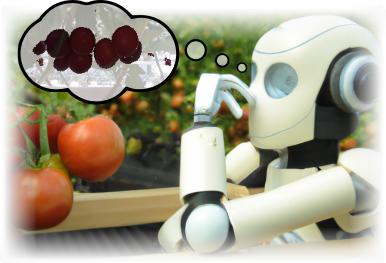}\vspace{-2mm}
\end{center}
\vspace{-5pt}
\caption{We propose \textbf{TomatoDIFF}, a diffusion--based model for semantic segmentation of on--plant tomatoes and introduce a new large--scale greenhouse dataset \textbf{Tomatopia}. The visualization shows TomatoDIFF results of segmented crops from our dataset. The image depicting a thinking robot was generated by the diffusion model DALL-E~2~\cite{dalle2_Ramesh_2022}. \label{fig:teaser}\vspace{-15pt}}
\end{figure}

Another major drawback of deep learning approaches is their need for massive amounts of training data~\cite{Tian_20201_cv_agriculture}. While pretraining models on commonly used large-scale datasets can partially alleviate this issue, fine-tuning the models still requires a moderately large set of precisely annotated images in a pixel-wise manner, to bridge the domain gap. To the best of our knowledge, currently there are very few tomato datasets that can be used for this purpose, as experiments in existing papers are predominately conducted on private databases. This lack of publicly available data not only hinders comparability and reproducibility of results but also negatively impacts benchmarking. 



In this paper we address all aforementioned problems and make the following contributions: \textit{i)} we propose \textbf{TomatoDIFF} -- a novel method for accurate semantic segmentation of on-plant tomatoes, based on denoising diffusion, \textit{ii)} we introduce \textbf{Tomatopia} -- a new, large--scale dataset of four different greenhouse tomato cultivars, manually annotated in a pixel--wise manner, \textit{iii)} we conduct a comprehensive evaluation of our method and compare it against different state--of--the--art (SOTA) models.





\begin{table*}[t]
\caption{Detailed description of publicly available tomato datasets and our newly collected Tomatopia. 
\vspace{-30pt}}
\label{tab:rw_datasets}
\begin{center}
\resizebox{\linewidth}{!}{%
\begin{tabular}{|l | c c c c c c|} 
 \hline
 \textbf{Name} & \textbf{Number of images} & \textbf{Modality} & \textbf{Size (px)} & \textbf{Annotations*} & \textbf{Tomato types} & \textbf{Production technology} \\ 
 \hline\hline
 AgRob & $449$ & RGB & $1280\times720$ & bb & plum & soil--based \\ 
 \hline
 LaboroTomato & $804$ & RGB & $3024\times4032$ & bb+pw & truss, cherry & soil--based \\ 
 \hline
 Rob2Pheno & $123$ & RGB-D & $1280\times720$ & bb+pw & truss & hydroponics \\
 \hline
 Tomato dataset & $895$ & RGB & $400\times500$ & bb & truss, cherry, plum & hydroponics \\ \hline
  \textbf{Tomatopia}[Ours] & $1080$ & RGB-D & $1280\times720$ & pw & truss, cherry, plum & hydroponics \\
  \hline
\end{tabular}}
\end{center}
\vspace{-3mm}
\hspace{10pt} \footnotesize * bb=bounding box, pw=pixel--wise
\vspace{-15pt}
\end{table*}
\vspace{-2pt}
\section{Related work}
\vspace{-2pt}

\textbf{Fruit detection and segmentation.} Early algorithms for fruit detection and segmentation rely on traditional approaches, that try to identify fruits with prior knowledge of their color and shape. Malik \textit{et al.}~\cite{Malik2019_seg_loc_rgbd} for instance propose semantic segmentation of tomatoes in the HSV color space. Once localized, individual fruits are separated with the watershed algorithm. Verma \textit{et al.}~\cite{Verma204_shape_tomato_seg} on the other hand propose calculating image gradients and approximating tomatoes with ellipses. In a follow--up paper~\cite{Verma2015_shape_tomato_seg_seq} the robustness of the system is increased by encompassing a temporal component, to track fruit progress over time. Point clouds acquired with depth cameras are also commonly used for geometric modeling of the fruits 
in the 3D space~\cite{Fu_2020_fruit_detection_review}. 

More recent applications, leverage different deep learning methods, that learn the characteristics of the fruits without any prior knowledge. Faster R-CNN was for example selected as a tomato detector by both, Mu \textit{et al.}~\cite{Mu_2020_faster_Rcnn} and Seo \textit{et al.}~\cite{Seo_2021_fastRcnn} for its robustness and fast data processing. Mask R-CNN is another commonly used method that in addition to the detection task also performs pixel--wise instance segmentation. While Afonso \textit{et al.}~\cite{Afonso_2020_maskRcnn_Rob2Pheno} and \textit{Zu et al.}~\cite{Zu_2021_maskrcnn_green_tomato} both use Mask R-CNN as an out--of--the--box algorithm, Xu \textit{et al.}~\cite{Xu_2022_depth_maskRcnn} propose utilizing depth images in addition to standard RGB input. Fonteijn \textit{et al.}~\cite{Fonteijn_2021_rob2pheno} further upgrade the model by replacing standard CNN backbones with a transformer. Differently, Yuan \textit{et al.}~\cite{Yuan_2020_cherry_SSD} propose Single Shot multi-box Detector (SSD), which is also used in~\cite{Tenorio_2021_tomato_maturity, Moreira_2022_agrob_rpiTomato, Magalhaes_2021_SSD_YOLO_agRob}. In addition to SSD, Magalhaes \textit{et al.}~\cite{Magalhaes_2021_SSD_YOLO_agRob} and Moreira \textit{et al.}~\cite{Moreira_2022_agrob_rpiTomato} also considered YOLO and found out that it achieves very similar results, but has slightly longer inference time. 

\textbf{Denoising Diffusion Models.} Recently Denoising Diffusion Models (DDMs) have emerged as highly effective methods for a variety of computer vision tasks~\cite{croitoru2022diffusion, Vishal2023_T2V_DDPM}. First introduced by Ho \textit{et al.}~\cite{Ho_DDPM_2020}, DDMs were shown to be capable of producing high-quality images from pure Gaussian noise. 
In a recent publication by Saharia \textit{et al.}~\cite{Saharia2022_palette}, diffusion models were successfully modified to perform other challenging 
tasks as well, including colorization, inpainting, uncropping, and JPEG restoration. 

This paper has also inspired adaptations of the DDMs for the purpose of semantic segmentation in image data. Amit \textit{et al.}~\cite{Amit_2021_segdiff} for instance proposed SegDiff, a DDM conditioned on RGB input images, that learns to iteratively refine the segmentation map by applying the diffusion denoising technique. Wu \textit{et al.}~\cite{Wu_2022_MedSegDiff} observe that static image as the condition is hard to learn and propose dynamic conditional encoding for each step. In a follow-up paper~\cite{Wu_2023_MedSegDiff_v2}, they also upgrade the architecture of their model MedSegDiff, by incorporating a transformer-based conditional UNet framework. Wolleb \textit{et al.}~\cite{wolleb_swiss_army_knife_2022} take a different approach, that guides the diffusion model by an external gradient. 

\textbf{Tomato datasets.} To the best of our knowledge, there are only four publicly available datasets of greenhouse tomatoes, i.e., AgRob~\cite{Moreira_2022_agrob_rpiTomato}, LaboroTomato~\cite{LaboroTomato}, Rob2Pheno~\cite{Afonso_2020_maskRcnn_Rob2Pheno} and Tomato dataset~\cite{Tomato_dataset_kaggle}. All four provide bounding box annotations for tomato detection, while ground truth masks for tomato segmentation are available only in LaboroTomato and Rob2Pheno. These two however predominantly represent non--occluded tomatoes in environments with sparse vegetation. Therefore, they do not reflect unconventional
production systems that are resource efficient, use less space and produce higher yields. More details about individual datasets are given in Table~\ref{tab:rw_datasets}.
\vspace{-2pt}
\section{Methodology}
\vspace{-2pt}
Here, we propose a novel method, TomatoDIFF, for semantic segmentation of tomatoes in an end--to--end manner, without prior crop detection. 

\begin{figure}[t]
\begin{center}
\centering
  \includegraphics[width=\linewidth]{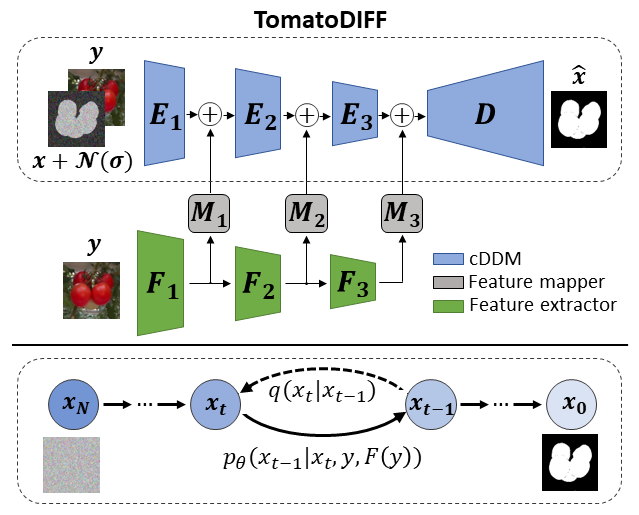}\vspace{-2mm}
\end{center}
\vspace{-5pt}
\caption{TomatoDIFF is trained with the denoising diffusion technique (bottom) and takes the RGB image, the noisy ground truth mask and the pretrained image features as input (top). In the test phase, the ground truth mask is substituted for pure 
Gaussian noise.\label{fig:DIFFrentTomatoes}\vspace{-15pt}}
\end{figure}

\textbf{Overview of TomatoDIFF.} A high-level overview of the proposed model  is presented in Figure~\ref{fig:DIFFrentTomatoes}. The backbone of TomatoDIFF represents a \textit{conditional} UNet trained with the recently emerging \textit{denoising diffusion} technique. It uses three different data sources, i.e. RGB tomato images, ground truth masks and feature maps from three different levels of a pretrained convolutional network $F$. Embeddings produced by $F$ are designed to improve the segmentation performance of the UNet by providing additional semantic characteristics of objects of interest. These pre-extracted maps are fed into intermediate layers of the encoder $E$ through a feature mapper $M$, which represents a convolutional module, that allows the autoencoder to select the most informative features. The training objective of TomatoDIFF is to generate a segmentation mask of tomatoes $\hat{x}$ by iteratively denoising the noisy input ground truth mask $x+\mathcal{N}(\sigma)$. The model is conditioned on the RGB image, denoted as $y$. Once TomatoDIFF is trained, the noisy ground truth mask is substituted for pure Gaussian noise. 

\textbf{Conditional Denoising Diffusion Models (cDDMs)} are conditional likelihood-based generative methods, designed to learn given data distribution $p_{data}(\mathbf{x})$ with standard deviation $\sigma_{data}$, by employing a two-stage approach, proposed by Ho \textit{et al.}~\cite{Ho_DDPM_2020}. In the first stage, a forward diffusion process is used to corrupt the data $\mathbf{x}_0\sim p_{data}(\mathbf{x})$ by gradually adding Gaussian noise $\mathcal{N}(0, \sigma^2\mathbf{I})$ to the sample $x_0$. This process produces a noisy sample $x_N$ and represents a non--homogeneous Markov chain described by the equation:
\begin{equation}\label{eq:forward}
    q(\mathbf{x}_t|\mathbf{x}_{t-1}) = \mathcal{N}(\mathbf{x}_t|\mathbf{x}_{t-1}\sqrt{1-\beta_t}, \beta_t\mathbf{I}),
\end{equation}
where $t$ is the time step from a predefined time sequence $\{t_0, t_1,... t_N\}$, while $\beta_t=\sigma_t^2$ defines the amount of noise added at each step. The value of $\beta_t$ is determined by a linear variance schedule, found to work best in terms of sampling speed and generated data quality~\cite{Karras2022edm}. 

In the second stage, a generative model parametrized by $\theta$ performs sequential denoising of $\mathbf{x}_N$, conditioned on $\mathbf{y}$:
\begin{equation}\label{eq:reverse}
    p_{\theta}(\mathbf{x}_{t-1}|\mathbf{x}_t, \mathbf{x}_0, \mathbf{y})=\mathcal{N}(\mathbf{x}_{t-1}|\tilde{\mu}_{t}(\mathbf{x}_t, \mathbf{x}_0), \tilde{\beta_t}\mathbf{I}),
\end{equation}
where $t=t_N, t_{N-1},... t_0$, $\tilde{\mu}_{t}(\mathbf{x}_t, \mathbf{x}_0)=\frac{\sqrt{\bar{\alpha}_{t-1}}\beta_t}{1-\bar{\alpha}_t}\mathbf{x}_0+\frac{\sqrt{\bar{\alpha}_t}(1-\bar{\alpha}_{t-1})}{1-\bar{\alpha}_t}x_t$ and $\tilde{\beta_t}=\frac{1-\bar{\alpha}_{t-1}}{1-\bar{\alpha}_{t}}\beta_t$. The mean function $\tilde{\mu}_{t}$ is optimized by an aproximator $\mathcal{D}_\theta(\mathbf{x}, \sigma)$, trained to minimize the expected $L_2$ denoising error:
\begin{equation}
    \mathcal{L} = \mathbb{E}_{\mathbf{x}\sim p_{data}}\mathbb{E}_{\mathbf{n}\sim \mathcal{N}(0, \sigma^2\mathbf{I})} ||\mathcal{D}(\mathbf{x}+\mathbf{n}; \sigma)-\mathbf{x}||_2^2,
\end{equation}
where $\mathcal{D}_\theta(\mathbf{x}, \sigma)$ is a neural network. In TomatoDIFF, an unconditional U-Net~\cite{PixelCNN_ICLR2017_Kingma} architecture, originally proposed in~\cite{Ho_DDPM_2020}, is selected for the implementation of this network. For efficiency reasons, we leverage the recently published DPM-Solver~\cite{Lu_NIPS2022_DPM_solver}, a dedicated high-order
solver for diffusion ordinary differential equations (ODEs).

\vspace{-2pt}
\section{Experimental Setup}
\vspace{-2pt}
\textbf{Datasets.} TomatoDIFF and competitive models are trained from scratch, on LaboroTomato and our dataset Tomatopia (Table~\ref{tab:rw_datasets}). In experiments conducted on LaboroTomato, we use the predetermined train/test split and resize all images to $512\times768$ pixels.

\begin{figure}[t]
\begin{center}
\centering
  \includegraphics[width=\linewidth]{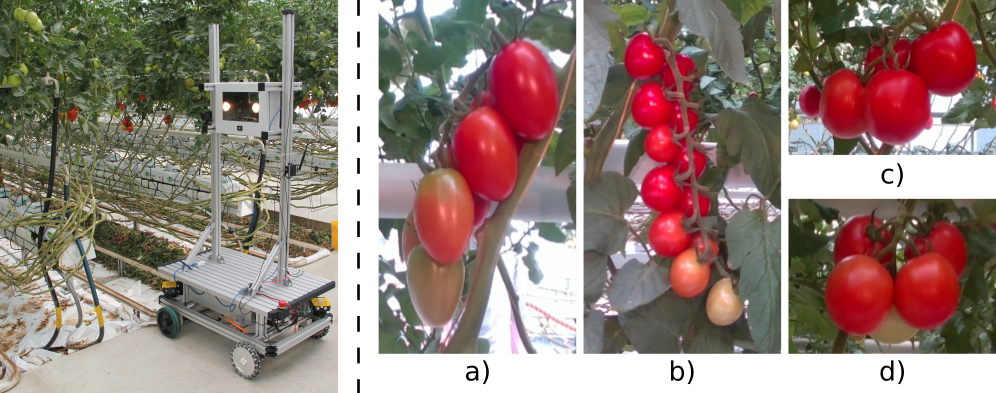}\vspace{-2mm}
\end{center}
\vspace{-3pt}
\caption{Image of the robot used for image acquisition (left) and tomato cultivars from Tomatopia (right): a) Ardiles (mini plum), b) DRC 564 (red cherry), c) Altabella (middle truss), d) Brilliant (large truss). \label{fig:tomato_cultivars} \vspace{-15pt}}
\end{figure}

\begin{figure}[h]
\begin{center}
\centering
  \includegraphics[width=\linewidth]{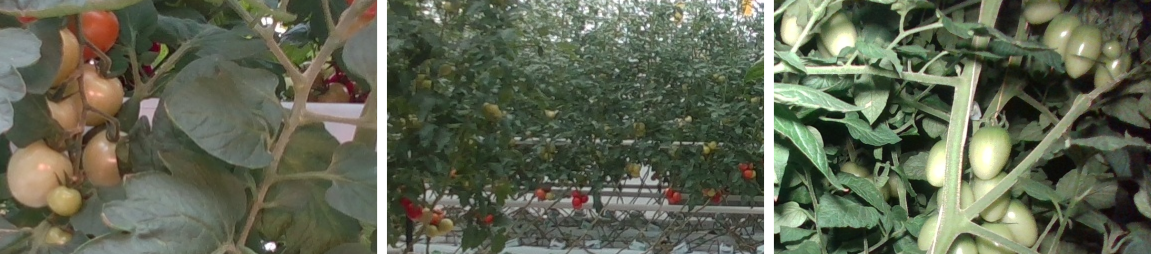}\vspace{-2mm}
\end{center}
\vspace{-3pt}
\caption{Examples of challenging scenes from Tomatopia: highly occluded and distant crops in daylight and night settings. \label{fig:challenging_scenes} \vspace{-15pt}}
\end{figure}

Tomatopia RGB images are resized to $768\times512$ pixels. Unlike existing, publicly available datasets, our image data is collected in an 
unconventional production environment that 
uses less space and produces higher yields. Tomato fruits are hence densely distributed and highly occluded (Figure~\ref{fig:challenging_scenes}). 
The average tomato-to-background ratio is 20\%. Tomatopia contains $4$ equally distributed tomato cultivars at various maturity levels, i.e. mini plum tomatoes Ardiles, red cherry tomatoes DRC 564, middle sized truss tomatoes Altabella and large truss tomatoes Brilliant~\ref{fig:tomato_cultivars}. The images were acquired with Intel RealSense D435I camera, attached to a robot that moves on rails installed in the middle of equally distanced rows with tomato plants (Figure~\ref{fig:tomato_cultivars}). Tomatopia consists of 
$612$ images taken at daylight and $466$ captured at night, using $2$ halogen lamps with custom polycarbonate diffusors, each with a power of $25$W. 
The density of the 
foliage varies from very dense to sparse. 

\textbf{Implementation details.} Apart from the RGB input image and the ground thruth mask, TomatoDIFF also leverages image features extracted at three different network levels of WideResNet50 \cite{WideResNet} (model $F$), pretrained on ImageNet. 
TomatoDIFF is optimized using AdamW~\cite{AdamW_ICLR2019}, with learning rate set to $0.0001$, weight decay of $0.001$ and $\beta_1$ and $\beta_2$ set to $0.95$ and $0.999$. The code is implemented in Python $3.8$ using PyTorch $1.9$ and CUDA $11.7$. Experiments are run on a single NVIDIA GeForce RTX $3090$, where TomatoDIFF required around $1.53 s$ to perform the segmentation of a single image, once the model is trained. The number of sampling iterations was set to $5$, experimentally found to speed up the algorithm, without result degradation. To optimize the size of the model, we split each image into $24$ non-overlapping patches of size $64\times64$, prior to the processing. Trained models are quantitatively evaluated in terms of Intersection-over-Union (IoU) as well as F1 score to 
account for the class imbalance between tomato and background pixels.
\vspace{-2pt}
\section{Results}\label{Sec: Results}
\vspace{-2pt}
\textbf{Quantitative analysis.} First, we compare TomatoDIFF against instance segmentation methods MaskRCNN~\cite{he_ICCV_2017_maskrcnn}, YOLACT~\cite{Bolya_2019_ICCV_yolact} and SOLOv2~\cite{Wang_NIPS2020_solov2}. We use only LaboroTomato images, since Tomatopia doesn't include bounding box labels for each tomato fruit. 
During the evaluation of competitive models, individual segmented instances are merged into one prediction map and compared with the ground truth mask of the whole image. Results are shown in Table~\ref{tab:quantitative_results_two_stage}. TomatoDIFF beats all three tested methods, outperforming the runner-up MaksRCNN for $4\%$ in terms of IoU and for $6\%$ in terms of F1 score.


\begin{table}[t]
\caption{Comparison of TomatoDIFF and commonly used instance segmentation models for fruit yield prediction. 
\vspace{-15pt}}
\label{tab:quantitative_results_two_stage}
\smallskip
\begin{center}
\resizebox{0.99\columnwidth}{!}{%
\def\arraystretch{1.5}
\begin{tabular}{ c l c | c | c | c  }
\toprule
 \multicolumn{2}{c|}{\multirow{2}{*}{\diagbox[innerwidth=75pt]{\textbf{Dataset}}{\textbf{Model}}}} & MaskRCNN & YOLACT & SOLOv2 & \textbf{TomatoDIFF} \\
 & \multicolumn{1}{c|}{}& \cite{he_ICCV_2017_maskrcnn} & \cite{Bolya_2019_ICCV_yolact} & \cite{Wang_NIPS2020_solov2} & [Ours]\\ \hline \hline 
 \multicolumn{1}{c|}{LaboroTomato} & \multicolumn{1}{c|}{\textbf{IoU}} & $79.34$ & $66.90$ & $74.63$ & $\mathbf{84.04}$ \\ \cline{2-6} 
 \multicolumn{1}{c|}{\cite{LaboroTomato}} & \multicolumn{1}{c|}{\textbf{F1}} & $84.14$ & $78.73$ & $81.23$ & $\mathbf{90.22}$ \\
\bottomrule
\vspace{-10pt}
\end{tabular}}
\end{center}
\end{table}

We also compare TomatoDIFF against commonly used models for semantic segmentation of crops. Results are shown in Table~\ref{tab:quantitative_results_semantic_segmentation}. Our model again shows best results in terms of IoU and F1 score. TomatoDIFF segmentation accuracy is also found to be more consistent accross both datasets, LaboroTomato and Tomatopia. SegFormer~\cite{Xie_NIPS_2021_segformer} achieves very competitive results on LaboroTomato, but finds the segmentation of Tomatopia tomatoes very challenging. BiSeNet v2 on the other hand is the runner-up in Tomatopia experiments, but frequently fails in the segmentation performed on LaboroTomato. In general, transformer-based models such as SegFormer are found to have difficulties segmenting highly occluded tomatoes from Tomatopia. At the same time, they seem to better learn the semantics of the non--occluded crops, even in datasets of tomatoes with non-consistent backgrounds.

\begin{table}[t]
\vspace{-12pt}
\caption{Comparison of TomatoDIFF and models for semantic segmentation of crops. 
\vspace{-15pt}} 

\label{tab:quantitative_results_semantic_segmentation}
\smallskip
\begin{center}
\resizebox{0.99\columnwidth}{!}{%
\def\arraystretch{1.5}
\begin{tabular}{  l | c | c | c | c}
\toprule
  \multirow{2}{*}{\diagbox[innerwidth=68pt]{\textbf{Model}}{\textbf{Dataset}}} &\multicolumn{2}{c|}{LaboroTomato~\cite{LaboroTomato}} & \multicolumn{2}{c}{Tomatopia[Ours]}\\ \cline{2-5}
  & \textbf{IoU} & \textbf{F1} & \textbf{IoU} & \textbf{F1} \\ \hline\hline
  UNet~\cite{Ronneberger_2015_UNet} & $78.92$ & $40.53$ & $78.50$ & $75.86$ \\ \midrule
  FCN-ResNet~\cite{Long_2015_CVPR_fcnresnet} & $79.70$ & $87.69$ & $75.04$ & $85.30$ \\ \midrule
  PSPNet~\cite{Zhao_2017_pspnet} & $74.14$ & $83.87$ & $74.44$ & $84.93$\\\midrule 
  DeepLab v3+~\cite{Chen_2018_ECCV_deeplabv3} & $64.68$ & $76.42$ & $69.44$ & $81.33$\\\midrule 
  PointRend~\cite{Kirillov_2020_CVPR_pointrend} & $73.88$ & $83.58$ & $76.82$ & $86.56$ \\ \midrule
  BiSeNet v2~\cite{Yu_2021_ICCV_bisenetv2} & $72.15$ & $82.52$ & $78.82$ & $86.88$ \\ \midrule
  SegFormer~\cite{Xie_NIPS_2021_segformer} & $82.02$ & $89.63$ & $72.00$ & $83.13$ \\ \midrule
  \textbf{TomatoDIFF} [Ours]& $\mathbf{84.04}$ & $\mathbf{90.22}$ & $\mathbf{81.25}$ & $\mathbf{88.64}$\\
\bottomrule
\end{tabular}}\vspace{-15pt}
\end{center}
\end{table}

\begin{figure}[h]
\begin{center}
\centering
  \includegraphics[width=\linewidth]{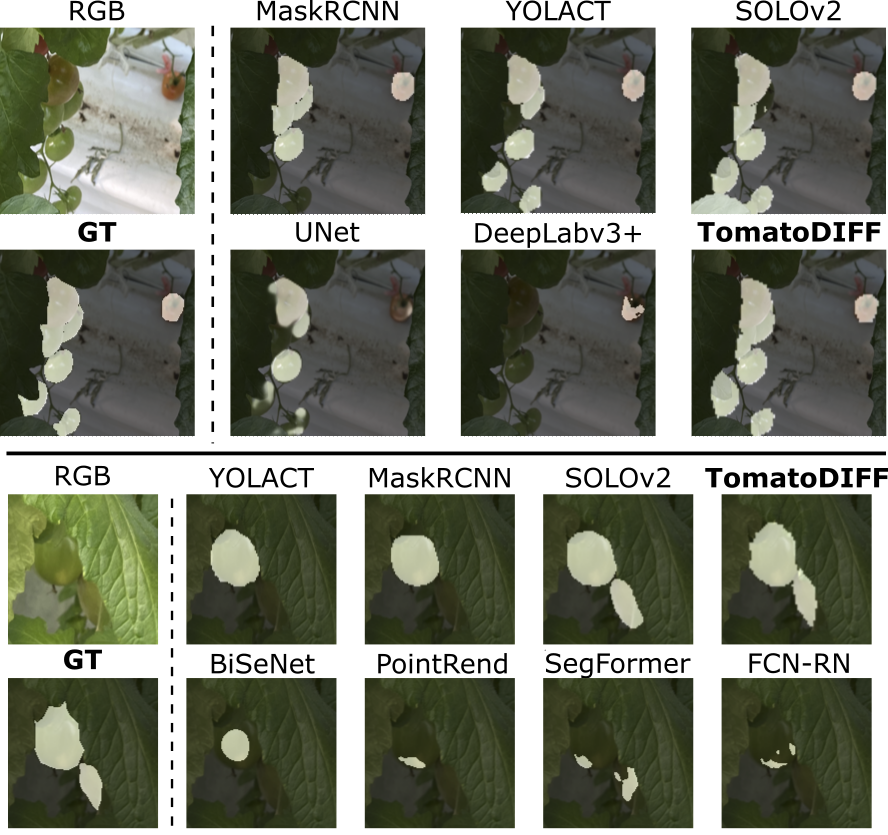}\vspace{-2mm}
\end{center}
\caption{TomatoDIFF shows most consistent result in terms of detection and segmentation of various tomatoes. Methods for semantic segmentation often fail to detect green crops, while methods for instance segmentation frequently fail to detect severely occluded fruits. \label{fig:selected_examples}\vspace{-12pt}}
\end{figure}

\begin{figure}[h]
\begin{center}
\centering
  \includegraphics[width=\linewidth]{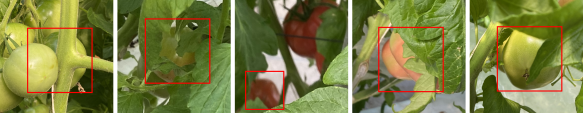}\vspace{-2mm}
\end{center}
\caption{Selected examples of tomatoes missed by TomatoDIFF (in red boxes). \label{fig:fails}\vspace{-12pt}}
\end{figure}

\textbf{Qualitative analysis.} 
Red tomatoes are almost always successfully detected by all tested models. Green tomatoes on the other hand are often left undetected by semantic segmentation models (Figure~\ref{fig:selected_examples}), while instance segmentation algorithms are overall 
more accurate in these scenarios. Nevertheless, all instance segmentation methods seem to have problems detecting severely occluded tomatoes. Interestingly, occluded crops are less challenging for TomatoDIFF, who learns the semantics of the fruits and is capable of predicting even hidden tomato parts. There are however some very challenging scenarios, where TomatoDIFF fails (Figure~\ref{fig:fails}). 

\textbf{Ablation study.} We investigate the contribution of key TomatoDIFF components in a study, where we test $3$ ablated models, i.e. A1) a cDDM model learned without pretrained feature maps and mapping modules $M_1, M_2, M_3$, A2) UNet with attention layers, trained without the denoising diffusion process, and A3) plain UNet. As can be seen in Table~\ref{tab:ablation_study}, TomatoDIFF outperforms all other models by a large margin. The diffusion model without integrated pretrained features gives worst results in terms of IoU and best in terms of F1 score. 
Non--diffusional variants on the other hand give far better IoU results, with A3 (UNet without attention layers) being just slightly worse than TomatoDIFF in terms of IoU score. Nevertheless, we note that this model produces a lot of false positives, due to background textures that are triggering incorrect model's response. Based on our analysis, we hypothesize that attention layers provide context and help the model to better learn the semantic characteristics of crops. Pretrained features on the other hand additionally enrich feature maps learned by the cDDM and improve the segmentation accuracy.

\begin{table}[h]
\caption{Results of the ablation study performed with 3 ablated models: A1) TomatoDIFF learned without pretrained features, A2) A1 trained without diffusion and A3) A2 without attention layers.}
\vspace{-15pt}
\label{tab:ablation_study}
\smallskip
\begin{center}
\resizebox{0.99\columnwidth}{!}{%
\def\arraystretch{1.5}
\begin{tabular}{  l | c | c | c | c}
\toprule
 \multirow{2}{*}{\diagbox[innerwidth=75pt]{\textbf{Model}}{\textbf{Dataset}}}&\multicolumn{2}{c|}{LaboroTomato~\cite{LaboroTomato}} & \multicolumn{2}{c}{Tomatopia[Ours]}\\ \cline{2-5}
 & \textbf{IoU} & \textbf{F1} & \textbf{IoU} & \textbf{F1} \\ \hline\hline
 \textbf{TomatoDIFF} & $\mathbf{84.04}$ & $\mathbf{90.22}$ & $\mathbf{81.25}$ & $\mathbf{88.64}$\\ \hline \hline
 A1: w/o $M_1, M_2, M_3$ & $64.05$ & $55.63$ & $65.63$ & $74.92$ \\ \midrule
 A2: A1 w/o diffusion & $70.72$ & $50.61$ & $72.66$ & $69.21$ \\ \midrule
 A3: A2 w/o att. layers & $78.92$ & $40.53$ & $78.50$ & $75.86$ \\ \bottomrule
\end{tabular}}\vspace{-15pt}
\end{center}
\end{table}
\vspace{-2pt}
\section{Conclusion}
\vspace{-3pt}
In this paper we propose a novel diffusion--based method TomatoDIFF for semantic segmentation of on--plant tomatoes. The model was evaluated and compared against different SOTA algorithms, achieving best overall results on two different datasets, i.e. LaboroTomato and Tomatopia, a newly collected dataset of greenhouse tomatoes representing an unconventional production environment with dense foliage. 

\bibliographystyle{ieee}
\bibliography{bibliography}





\end{document}